\documentclass[10pt,twocolumn,letterpaper]{article}

\usepackage{cvpr}
\usepackage{times}
\usepackage{epsfig}
\usepackage{graphicx}
\usepackage{amsmath}
\usepackage{amssymb}
\usepackage{mathtools}
\usepackage{multirow}
\usepackage{enumitem}
\usepackage{caption}
\usepackage{subcaption}
\usepackage{sidecap}

\usepackage[capbesideposition=outside,capbesidesep=quad]{floatrow}
\makeatletter
\renewcommand{\paragraph}{%
  \@startsection{paragraph}{4}%
  {\z@}{1ex \@plus 1ex \@minus .2ex}{-1em}%
  {\normalfont\normalsize\bfseries}%
}
\makeatother
\usepackage[pagebackref=true,breaklinks=true,letterpaper=true,colorlinks,bookmarks=false]{hyperref}
\newcommand{\q}[1]{``#1"}

\newcommand{\colskip}{\hskip 1cm}

\cvprfinalcopy 

\def\cvprPaperID{2459} 

\ifcvprfinal\fi
\begin{document}

\title{Handwriting Recognition in Low-resource Scripts using Adversarial Learning}

\author{Ayan Kumar Bhunia\textsuperscript{1} \hspace{.2cm} Abhirup Das\textsuperscript{2} \hspace{.2cm} Ankan Kumar Bhunia\textsuperscript{3}\hspace{.2cm}   Perla Sai Raj Kishore\textsuperscript{2}\hspace{.2cm} Partha P. Roy\textsuperscript{3} \\
\textsuperscript{1}Nanyang Technological University, Singapore \hspace{.1cm} \textsuperscript{2} Institute of Engineering \& Management, India 
\hspace{.1cm}  \\ \textsuperscript{3} Jadavpur University, India \hspace{.1cm} \textsuperscript{4} Indian Institute of Technology Roorkee, India\\
{\tt\small \textsuperscript{1}ayanbhunia@ntu.edu.sg }
}
\maketitle
\ifcvprfinal\thispagestyle{empty}\fi

\begin{abstract}
Handwritten Word Recognition and Spotting is a challenging field dealing with handwritten text possessing irregular and complex shapes. The design of deep neural network models makes it necessary to extend training datasets in order to introduce variations and increase the number of samples; word-retrieval is therefore very difficult in low-resource scripts. Much of the existing literature comprises preprocessing strategies which are seldom sufficient to cover all possible variations. We propose the Adversarial Feature Deformation Module (AFDM) that learns ways to elastically warp extracted features in a scalable manner. The AFDM is inserted between intermediate layers and trained alternatively with the original framework, boosting its capability to better learn highly informative features rather than trivial ones. We test our meta-framework, which is built on top of popular word-spotting and word-recognition frameworks and enhanced by the AFDM, not only on extensive Latin word datasets but also sparser Indic scripts. We record results for varying training data sizes, and observe that our enhanced network generalizes much better in the low-data regime; the overall word-error rates and mAP scores are observed to improve as well.
\end{abstract}


\section{Introduction}

Handwriting recognition has been a very popular area of research over the last two decades, owing to handwritten documents being a personal choice of communication for humans, other than speech. The technology is applicable in postal automation, bank cheque processing, digitization of handwritten documents, as well as reading aid for the visually handicapped. Handwritten character recognition and word spotting and recognition systems have evolved significantly over the years. Since Nipkow's scanner \cite{pal2004indian} and LeNet \cite{lecun1998gradient}, modern deep-learning based approaches today \cite{krishnan2016deep, poznanski2016cnn, sudholt2017evaluating} seek to be able to robustly recognize handwritten text by learning local invariant patterns across diverse handwriting styles that are consistent in individual characters and scripts. These deep learning algorithms require vast amounts of data to train models that are robust to real-world handwritten data. While large datasets of both word-level and separated handwritten characters are available for scripts like Latin, a large number of scripts with larger vocabularies have limited data, posing challenges in research in the areas of word-spotting and recognition in languages using these scripts.   

Deep learning algorithms, which have emerged in recent times, enable networks to effectively extract informative features from inputs and automatically generate transcriptions \cite{shi2017end} of images of handwritten text or spot \cite{sudholt2016phocnet} query words, at high accuracy. In the case of scripts where abundant training data is not available, DNNs often fall short, overfitting on the training set and thus poorly generalizing during evaluation. Popular methods such as data augmentation allow models to use the existing data more effectively, while batch-normalization \cite{Ioffe:2015:BNA:3045118.3045167} and dropout \cite{srivastava2014dropout} prevent overfitting. Augmentation strategies such as random translations, flips, rotations and addition of Gaussian noise to input samples are often used to extend the original dataset \cite{krizhevsky2012imagenet} and prove to be essential for not only limited but also large datasets like Imagenet \cite{deng2009imagenet}. The existing literature \cite{cirecsan2015multi, krishnan2018word, poznanski2016cnn, zhong2015high} augment the training data prior to feature extraction before classifying over as many as 3755 character classes \cite{zhong2015high}. Such transformations, however, fail to incorporate huge variation in writing style and the complex shapes assumed by characters in words, by virtue of the free-flowing nature of handwritten text. Due to the huge space of possible variances in real handwritten images, training by generating deformed examples through such generic means is not sufficient as the network easily adapts to these policies. Models need to become robust to uncommon deformations in inputs by learning to effectively utilize the more informative invariances, and it is not optimal to utilize just ``hard" examples to do so \cite{shrivastava2016training, takac2013mini}. Instead, we propose an adversarial-learning based framework for handwritten word retrieval tasks for low resource scripts in order to train deep networks from a limited number of samples. 

Information retrieval from handwritten images can be mainly classified into two types: (a) Handwritten Word Recognition (HWR) which outputs the complete transcription of the word-image and (b) Handwritten Word Spotting(HWS) which finds occurrences of a query keyword (either a string or sample word-image) from a collection of sample word-images. The existing literature of deep-learning based word retrieval, which cover mostly English words, make use of large available datasets, or use image augmentation techniques to increase the number of training samples \cite{krishnan2018word}. Bhunia \etal \cite{bhunia2018cross} proposed a cross-lingual framework for Indic scripts where training is performed using a script that is abundantly available and testing is done on the low-resource script using character-mapping. The feasibility of this approach mostly depends on the extent of similarity between source and target scripts.  
Antoniou \etal \cite{antoniou2017data} proposed a data augmentation framework using Generative Adversarial Networks (GANs) which can generate augmented data for new classes in a one-shot setup. 

Inspired by the recent success of adversarial learning for different tasks like cross-domain image translation \cite{CycleGAN2017}, domain adaptation \cite{tzeng2017adversarial} etc. we propose a generative adversarial learning based paradigm to augment the word images in a high dimensional feature space using spatial transformations \cite{jaderberg2015spatial}. We term it as the Adversarial Feature Deformation Module that is added on top of the original task network performing either recognition or spotting. It prevents the latter from overfitting to easily learnable and trivial features. Consequently, frameworks enhanced by the proposed module generalize well to realistic testing data with rare deformations. Both the adversarial generator(AFDM) and task network are trained jointly where adversarial generator intends to create \q{hard} examples while the task network will attempt to learn invariances to difficult variations, gradually becoming better over time.  In this paper, we make the following novel contributions: 
\begin{enumerate}[wide, labelwidth=!, labelindent=0pt, nosep, topsep=0pt]
\item We propose a scalable solution to HWR and HWS in low resource scripts using adversarial learning to augment the data in high-dimensional convolutional feature space. Various deformations introduced by the adversarial generator encourage the task network to learn from different variations of handwriting even from a limited amount of data. 
\item We compare our adversarial augmentation method with different baselines, and it clearly shows that the proposed framework can improve the performance of state-of-the-art handwritten word spotting and recognition systems. Not only is the performance improved in the case of low-resource scripts, but models generalize better to real-world handwritten data as well. 
\end{enumerate}

\section{Handwritten Word Retrieval Models} 

We use the CRNN \cite{shi2017end} and PHOCNet \cite{sudholt2016phocnet} as the baseline framework for handwritten word recognition and spotting respectively; on top of these, we implement our adversarial augmentation method. Significantly, our model is a meta-framework in the sense that the augmentation module can be incorporated along with a ResNet-like architecture too, instead of the VGG-like architecture adopted originally in both frameworks. 

\paragraph{Convolutional Recurrent Neural Network for HWR:} Shi \etal \cite{shi2017end} introduced an end-to-end trainable Convolutional Recurrent Neural Network with Connectionist Temporal Classification (CTC) loss which can handle word sequences of arbitrary length without character segmentation and can predict the transcription of out-of-vocabulary word images using both lexicon-based and lexicon free approaches. 
The `Map-to-Sequence' layer \cite{shi2017end} acts as the bridge between the convolutional and the recurrent layers. The input is first fed to the convolutional layers; a recurrent network is built to make a per-frame prediction for each frame of the extracted features. Finally, a transcription layer translates the prediction from the recurrent layers into a label sequence.
 
\paragraph{PHOCNet for HWS:}The PHOCNet \cite{sudholt2016phocnet} is a state-of-the-art approach in word-spotting, achieving exemplary results for both QbE (Query by Example) and QbS (Query by String) methods. The model reduces images of handwritten words to encoded representations of their corresponding visual attributes. The PHOC label \cite{sudholt2016phocnet} of a word is obtained by segmenting it into histograms at multiple levels. The histograms of characters in a word and its n-grams are calculated and concatenated to obtain a final representation. Once trained, an estimated PHOC representation is predicted for input word-images of varying sizes, by using a Spatial Pyramid Pooling layer \cite{he2014spatial}. These semantic representations of query and word-images can be compared directly by simple nearest-neighbor search (for QbE) or compared with the output representation of the deep model with PHOC of word-images in the dataset (for QbS). The PHOCNet uses sigmoid activation to generate the histograms instead of Softmax, utilizing a multi-label classification approach. 

\section{Related Works} 

Handwriting recognition has been researched in great detail in the past and in-depth reviews exist about it \cite{plamondon2000online}. Nevertheless, the search for a better and more accurate technique continues to date. Results presented in \cite{jaderberg2014synthetic} show that models should preferably use word-embeddings over bag-of-n-grams approaches. Based on this, another approach \cite{poznanski2016cnn} employed a ConvNet to estimate a frequency based profile of n-grams constituting spatial parts of the word in input images and correlated it with profiles of existing words in a dictionary, demonstrating an attribute-based word-encoding scheme. In \cite{sudholt2016phocnet}, Sudholt \etal adopted the VGG-Net \cite{simonyan2014very} and used the terminal fully connected layers to predict holistic representations of handwritten-words in images by embedding their pyramidal histogram of characters (PHOC\cite{almazan2014word}) attributes. Architectures such as \cite{krishnan2016deep, sudholt2016phocnet, wilkinson2016semantic} similarly embedded features into a textual embedding space. The paper \cite{wilkinson2017neural} demonstrated a region-proposal network driven word-spotting mechanism, where the end-to-end model encodes regional features into a distributed word-embedding space, where searches are performed. Sequence discriminative training based on Connectionist Temporal Classification (CTC) criterion, proposed by Graves \etal in \cite{graves2006connectionist} for training RNNs \cite{hochreiter1997long} has attracted much attention and been widely used in works like \cite{graves2009novel, shi2017end}. In Shi \etal \cite{shi2017end}, the sequence of image features engineered by the ConvNet is given to a recurrent network such as LSTM \cite{graves2009novel} or MDLSTM \cite{voigtlaender2016handwriting, bluche2017scan} for computing word transcriptions. Authors in \cite{krishnan2018word} additionally included an affine-transformation based attention mechanism to reorient original images spatially prior to sequence-to-sequence transcription for improved detection accuracy. In most of the aforementioned methods, it is important to preprocess images in different ways to extend the original dataset, as observed in \cite{krishnan2016deep, krishnan2018word, krizhevsky2012imagenet, poznanski2016cnn, simard2003best}.

The process of augmenting to extend datasets is seen even in the case of large extensive datasets \cite{krishnan2018word, deng2009imagenet} and in works focusing on Chinese handwritten character recognition where there are close to 4000 classes in standard datasets. In a different class of approaches, the process of online hard example mining (OHEM) has proved effective, boosting accuracy in datasets by targeting the fewer \q{hard} examples in the dataset, as shown in  \cite{loshchilov2015online, shrivastava2016training, simo2014fracking, wang2015unsupervised}. With the advent of adversarial learning and GANs in recent years, several approaches have incorporated generative modeling to create synthetic data that is realistic \cite{dong2017unsupervised, odena2016conditional, yeh2016semantic}, following architectural guidelines described by Goodfellow \etal for stable GAN training \cite{goodfellow2014generative}. Papers such as \cite{antoniou2017data} use GANs to augment data in limited datasets by computing over a sample class image to output samples that belong to the same class.  

A recent work by Wang \etal \cite{wang2017fast} describes an adversarial model that generates hard examples by using the generator \cite{goodfellow2014generative} to incorporate occlusions as well as spatial deformations into the feature space, forcing the detector to adapt to uncommon and rare deformations in actual inputs to the model. In our framework, we use a similar strategy to make our word-retrieval detector robust and invariant to all sorts of variations seen in natural images of handwritten text. 
Another similar approach \cite{song2018vital} also explores the use of adversarial learning in visual tracking and detection of objects and attempts to alleviate the class-imbalance problem in datasets, where it is observed that the amount of data in one class far exceeds another class. Having a larger number of easy to recognize samples in datasets deters the training process as the detector is unaware of more valuable ``hard" examples.

\begin{figure*}[t]
\begin{center}
\includegraphics[width=0.95\linewidth]{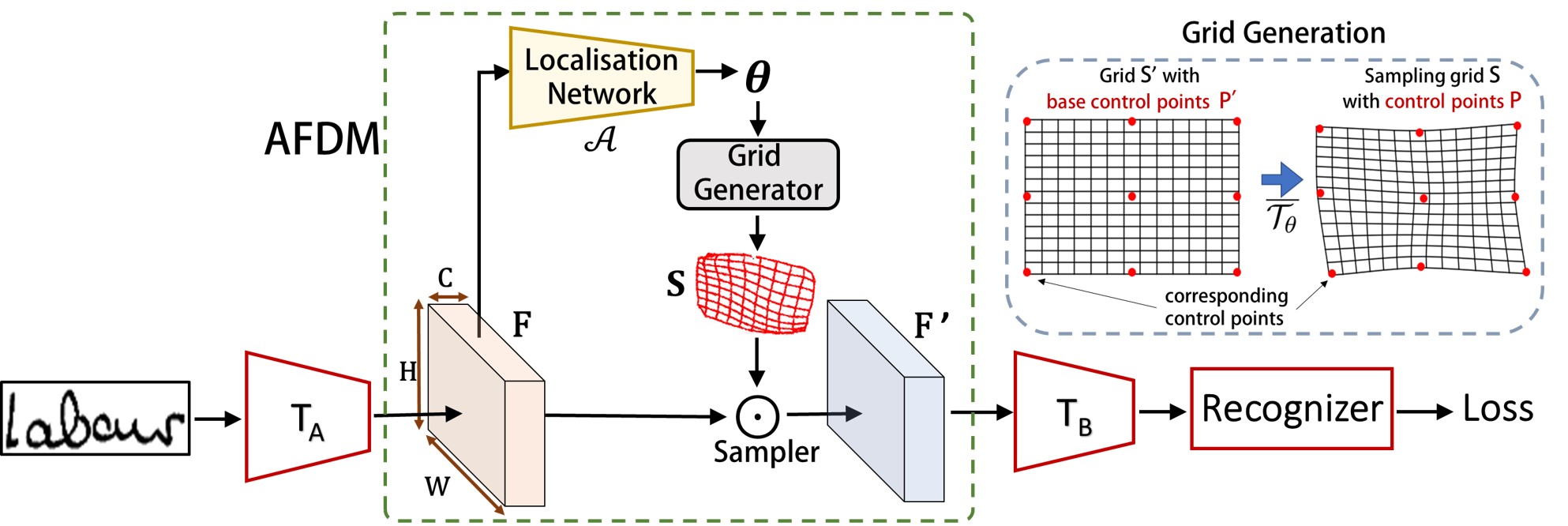}
\end{center}
   \caption{The architecture of our training network with the Adversarial Feature Deformation Module including the Localisation Network, Grid Generator and the Sampler inserted in between $T_A$ and $T_B$ of the task-network. The illustration depicts the use of the AFDM to uniformly deform the complete feature map $\mathbf{F}$.}
\label{fig:short}
\end{figure*}

\section{Proposed Methodology}

\subsection{Overview}

The generic augmentation techniques popularly observed in HWR and HWS frameworks are often insufficient for models to generalize to real handwritten data, especially in the case of low-resource scripts where existing datasets are small and cover only a fraction of irregularities observed in the real world. We propose a modular deformation network that is trained to learn a manifold of parameters seeking to deform the features learned by the original task network, thus encouraging it to adapt to difficult examples and uncommon irregularities.   

Let $T$ be the task network whose input is an image $I$. By task network, we mean either word recognition network \cite{shi2017end} or word spotting network \cite{sudholt2016phocnet}, and the corresponding task loss be $L_{task}$ which can be either CTC loss \cite{shi2017end} (for word recognition) or cross-entropy loss \cite{sudholt2016phocnet} (for word spotting); we will use the terms \textit{task network} and \textit{task loss} for simplicity of description. We introduce an  Adversarial Feature Deformation Module (AFDM) after one of the intermediate layers of the task network. Let us consider the task network $T$ to be dissected into three parts, namely $T_{A}$, $T_{B}$ and $R$. $R$ is the final label prediction part, predicting either the word-level transcription for recognition or PHOC labels \cite{sudholt2016phocnet} for word spotting; $T_{A}$ and $T_{B}$ are the two successive convolutional parts of task network $T$. The exact position of dissection between $T_{A}$ and $T_{B}$ is discussed in Section 5.1. Let us assume $F$ is the output feature map of $T_{A}$, i.e. $\mathbf{F} = T_{A}(I)$. The warped feature-map $\mathbf{F}'$ by AFDM is thereafter passed through $T_{B}$ and $R$ for final label prediction. While the complete task network $T$ is trained with the objective to correctly predict the output label, the AFDM tries to deform the features so that $T$ can not predict the correct label easily. $T$ is thereby enforced to generalize better to more discriminative invariances and informative features in the handwritten text data. The feature deformation network $\mathcal{A}$ and task network $T$ compete in this adversarial game while training. During inference, we only use $T$.

\subsection{Adversarial Feature Deformation Module}

The AFDM, inspired by Spatial Transformation Networks (STN) \cite{jaderberg2015spatial}, fulfills our objective of warping the features learned by $T_A$ to make recognition (or spotting) difficult for the task-network. 
The module uses its adversarial \textit{Localisation Network} $\mathcal{A}$ to predict a set of parameters $\theta$. These parameters are needed to compute the transformation matrix $\mathcal{T}_\theta$. The \textit{Grid Generator} generates a sampling grid $S$ by performing the transformation $\mathcal{T}_\theta$ on points in the grid $S'$ representing coordinates in $\mathbf{F'}$. The obtained grid $S$ represents $N$ points in the original map $\mathbf{F}$ where corresponding points in the target map $\mathbf{F'}$ should be sampled from such that the latter appears spatially warped in the manner described by $\mathcal{T}_\theta$. This grid $S$ and the original feature map are then given to the \textit{Bilinear Sampler} to obtain the target feature map $\mathbf{F}'$. Although a number of transformations \cite{jaderberg2015spatial} can be used in the AFDM, Thin Plate Spline Transformation (TPS) \cite{bookstein1989principal} is suggested to be the most powerful according to Jaderberg \etal \cite{jaderberg2015spatial}. 
We use TPS because of its degree of flexibility and ability to elastically deform a plane by solving a two-dimensional interpolation problem: the computation of a map $\mathbb{R}^2 \rightarrow \mathbb{R}^2$ from a set of arbitrary control points \cite{bookstein1989principal}. Furthermore, the matrix operations for grid-generation and transformation in TPS being differentiable, the module can backpropagate gradients as well.  

The parameters predicted adversarially by $\mathcal{A}$ denote $K$ control points $P = [p_1,\cdots, p_K] \in \mathbb{R}^{2\times K}$ with $p_v = [x_v, y_v]^T$ pointing to coordinates in $\mathbf{F}$ by regressing over their $x, y$ values, which are normalised to lie within [-1,1]. The Grid Generator uses the parameters representing the control points in $P$ to define a transformation function for a set of corresponding control points $P' = [p'_1,\cdots, p'_K]$, called the \textit{base control points} representing positions in $\mathbf{F}'$. Since the base control points are unchanging, $P'$ is a constant. 
The transformation is denoted by a matrix $\mathcal{T}_{\theta} \in \mathbb{R}^{2 \times (K+3)}$ that can be computed as:
\begin{equation}\label{T_tps}
\mathcal{T}_{\theta} = \begin{pmatrix} \Delta_{P'}^{-1} \begin{bmatrix} P^T \\0^{3\times2} \end{bmatrix} \end{pmatrix}^T
\end{equation} 
where $\Delta_{P'}\in \mathbb{R}^{(K+3)\times (K+3)}$ is also a constant. It is given by:
\begin{equation}
\Delta_{P'} = \begin{pmatrix} 1^{K \times 1} & P^{'T} & \mathcal{E} \\ 
                            0 & 0 & 1^{1 \times K} \\
                            0 & 0 & P' \end{pmatrix}
\end{equation}
where, the element in $i$-th row and $j$-th column of matrix $\mathcal{E}$ represents the euclidean distance between the base control points $p'_i$ and $p'_j$. 
Now, given that the grid of target points in $\mathbf{F'}$ is denoted as $S'=\{s'_i\}_{i=1,\cdots,N}$, with $s'_i=[\mathrm{x}'_i,\mathrm{y}'_i]^T$ being the x,y coordinates for the $i$-th point of a total of $N$ feature-points, for every point $s'_i$ we find the corresponding sampling position $s_i=[\mathrm{x}_i,\mathrm{y}_i]^T$ in $\mathbf{F}$ through the following steps:
\vspace{-0.05cm}
\begin{align}
&e'_{i,k} = d^2_{i,k} \ln d^2_{i,k}\\
&\hat{\mathrm{s}}'_i = [1,x'_i,y'_i,e'_{i,1},\cdots,e'_{i,K}]^T \in \mathbb{R}^{(K+3)\times 1}\\
&\mathrm{s}_i = \mathcal{T}_{\theta} \cdot \hat{\mathrm{s}}'_i\label{point_tps}
\end{align}

where $d_{i,k}$ is the euclidean distance between $s'_i$ and $k$-th base control point $p'_k$. We iteratively modify all $N$ points in $S'$ using eqn. (5) to define the grid-transform function $\overline{\mathcal{T}_{\theta}}(\cdot)$ and produce sampling grid $S$:
\begin{equation}\label{point_tps_last}
S = \overline{\mathcal{T}_{\theta}}(\{s'_i\}), \quad i=1,2,\cdots,N
\end{equation} 
We obtain the grid $S=\{s_i\}_{i=1,\cdots,N}$ representing sampling points in $\mathbf{F}$. 

The network represented by $\mathcal{A}$ includes a final fully-connected ($fc$) layer predicting 2K normalized coordinate values. It is fitted with the $\tanh(\cdot)$ activation function, after which the values are reshaped to form matrix the P. It is to be noted that the aforementioned equations define the deformation operation executed by the AFDM such that all channels in our original map are deformed uniformly. In the later sections, we discuss the partitioning strategy where smaller sub-maps in $\mathbf{F}$ are fed individually into it for deformation.

\subsection{Adversarial Learning} 

Traditional approaches to adversarial learning \cite{goodfellow2014generative} involve training a model to learn a generator $G$ which given a vector $z$ sampled from a noise distribution $P_{noise}(z)$ outputs an image $G(z)$. The discriminator $D$ takes either the generated image or real image $x$ from distribution $P_{data}(x)$ as input, and identifies whether it is real or fake. 
The objective function for training the network using cross-entropy loss is defined as:
\vspace{-0.1cm}
\begin{multline}\label{loss_0}
{L} = \min\limits_{G} \max\limits_{D} \hspace{0.1cm} \mathbb{E}_{x\sim P_{data}(x)}[log D(x)] \\
+ \mathbb{E}_{z\sim P_{noise}(z)}[log (1 - D(G(z)))]
\end{multline}
Adversarial learning trains $G$ to produce image statistics similar to that of training samples that the discriminator cannot distinguish, while training $D$ to declare $G(z)$ as fake; this is hardly accomplished by the objective functions used in supervised learning. In recognition based problems, $G$ is removed and $D$ is retained for inference.

Inspite of GANs being used in several computer vision tasks, direct use of eqn. \ref{loss_0} in HWR or HWS has not been seen. In the proposed model, the the generator framework $\mathcal{A}$ computes encoded features rather than random noise. Also, we seek to train the discriminating network, i.e. the task-network $T$ in a supervised manner using labelled samples, while encouraging it to accurately retrieve handwritten inputs despite deformations present in them.   

Now, instead of deforming $\mathbf{F}$ (with height H, width W and C channels) uniformly, we modify $k$ sub-maps constituting it in $k$ different ways ($k$ being a value much smaller than the number of channels $C$), thereby increasing the complexity of the task and preventing $\mathcal{A}$ from learning trivial warping strategies. $\mathbf{F}$ is divided into sub-maps $f_1$ through $f_k$, each of which has $\frac{C}{k}$ channels. The $m$-th sub-map $f_m \in \mathbb{R}^{H\times W\times \frac{C}{k}}$ is then fed into the $\mathcal{A}$ to generate $\theta_m$, and compute the grid-transform function 
$\overline{\mathcal{T}_{\theta_m}}(\cdot)$. The latter, as shown in eqn. \ref{T_tps} through \ref{point_tps_last}, transforms a given grid $S'_m$ to obtain the corresponding grid of sampling points $S_m$ for points belonging to sub-map $f_m$. The deformed feature map $\mathbf{F'}$ is thus computed as:
\begin{equation}
     \mathbf{F'} = (f_1\odot S_1) \hspace{0.1cm}\oplus\hspace{0.1cm} (f_2\odot S_2) \hspace{0.1cm}\oplus\hspace{0.1cm} \cdots \hspace{0.1cm}\oplus\hspace{0.1cm} (f_k\odot S_k) 
\end{equation}
where $\oplus$ denotes the channel-wise concatenation operation and $\odot$ denotes the bilinear-sampling mechanism corresponding to the transforms described in \cite{jaderberg2015spatial}. The sub-map $f_m$ is thus sampled to obtain $(f_m\odot S_m) \in \mathbb{R}^{H\times W\times \frac{C}{k}}$, and concatenated to get $\mathbf{F}'$ having same dimensions as the original feature-map. The AFDM thus learns a function $\mathcal{A}(\cdot)$ that computes the encoded features in the $m$-th sub-map to generate $\theta_m = \mathcal{A}(f_m)$. 

In absence of the AFDM (e.g. during testing), the output $\mathbf{F}$ of sub-network $T_{A}$  is further passed through $T_{B}$ and $R$. The recognizer $R$ outputs the predicted word-label $\mathrm{L}_p$ for a word image $I$. The word-label can be either word-level annotation represented by series of character, or PHOC label \cite{sudholt2016phocnet} based on the type of system. Let us assume the ground-truth label for the latter be $\mathrm{L}_g$. Thus our original word-retrieval loss $L_{task}$ can be defined as:
\begin{equation}\label{loss_1}
    L_{task} = Q_{\mathit{word}} (\mathrm{L}_p, \mathrm{L}_g) 
\end{equation}
where $Q_{\mathit{word}}(\cdot)$ represents a general function that computes loss between the prediction $\mathrm{L}_p$ and the ground truth label $\mathrm{L}_g$, which is either the CTC loss used in \cite{shi2017end} or the sigmoid-cross-entropy loss described in \cite{sudholt2016phocnet}.

During training, we have two different networks: the task network $T$ and Localisation Network $\mathcal{A}$. Let us consider their parameters to be $\theta _{T}$ and $\theta _{A}$ respectively. In one iteration during training, the data flow in the forward pass  is as follows: $I \rightarrow  T_{A}(\cdot) \rightarrow  AFDM(\cdot) \rightarrow T_{B}(\cdot) \rightarrow  R(\cdot)  \rightarrow  \mathrm{L}_p$, where $AFDM(\cdot)$ represents the complete deformation operation including parameter prediction by $\mathcal{A}$, grid-generation and sampling operations; the last two do not involve learning any parameters. $\mathcal{A}$  needs to learn feature deforming strategies through $\theta _{A}$  so that the recognizer should fail. We thus obtain $\theta _{A}$ by maximizing the loss function $L_{task}$. On the other hand, the $\theta _{T}$ is optimized to minimize the task loss $L_{task}$.
\begin{align}\label{loss_2}
\theta_{A} &= \arg \max\limits_{\theta_{A}} L_{task}  \\
\theta_{T} &= \arg \min\limits_{\theta_{T}} L_{task}  
\end{align}

As a result, if the deformation caused by the AFDM makes the ${I}$ hard to recognize, the task network $T$ gets a high loss and $\mathcal{A}$ gets a low loss, else if the modified features are easy to recognize, the $\mathcal{A}$ suffers a high loss instead. 


\begin{table*}
\label{table:tab1}
\centering
\caption{Performance of Handwritten Word Recognition(HWR) on different datasets}
\begin{tabular}{ccc@{\colskip}c@{\colskip}c@{\colskip}c@{\colskip}c@{\colskip}c@{\colskip}c@{\colskip}c}
\hline
\multirow{2}{*}{} & \multirow{2}{*}{Models} & \multicolumn{2}{c@{\colskip}}{IAM} & \multicolumn{2}{c@{\colskip}}{RIMES} & \multicolumn{2}{c@{\colskip}}{IndicBAN} & \multicolumn{2}{c}{IndicDEV} \\ 
\cline{3-10}     &                           & WER        & CER        & WER         & CER         & WER           & CER          & WER           & CER          \\ \hline
\multirow{5}{*}{\rotatebox[origin=c]{90}{Unconstrained}} 
                  & B1             &      23.14      &      12.02     &       16.04      &      11.17    &    26.31     &       14.67       &        25.35       &        13.69\\
                  & B2          &               \_&           \_&           \_&            \_ &    25.17     &       13.08       &        24.37       &        12.14\\
                  & B3             &      21.58      &      11.45      &       14.61      &      10.37     &         20.28&         11.13&         19.07&  10.34\\
                  & B4             &      19.97      &      9.81      &       12.42      &      7.61     &         17.67&          9.19&         16.46&   8.34\\
                  & \textbf{Ours}           &      \textbf{17.19}      &      \textbf{8.41}      &       \textbf{10.47}       &      \textbf{6.44}     &         \textbf{15.47}&          \textbf{7.12}&         \textbf{14.3} &   \textbf{6.14}\\ \hline
\multirow{5}{*}{\rotatebox[origin=c]{90}{Lexicon}} 
                  & B1             &       15.98&       10.05&        12.51&        9.64&          16.67&         10.21&          15.67&        9.78\\
                  & B2             &       \_&       \_&       \_&       \_&          15.87&         9.47&          14.69&        8.41\\
                  & B3             &       12.17&       8.45&        10.13&        7.17&          11.37&         7.64&          10.24&        6.76\\
                  & B4             &       10.24&       7.21&        7.59&        5.56&          9.69&         5.41&          8.67&        4.67\\
                  & \textbf{Ours}           &       \textbf{8.87}&       \textbf{5.94}&        \textbf{6.31}&        \textbf{3.17}&          \textbf{7.49}&         \textbf{4.37}&          \textbf{6.59}&    \textbf{3.97}\\         \hline
\end{tabular}
\end{table*}
\begin{table*}
\label{table:tab2}
\centering
\caption{Performance of Handwritten Word Spotting(HWS) on different datasets}
\begin{tabular}{cc@{\colskip}c@{\colskip}c@{\colskip}c@{\colskip}c@{\colskip}c@{\colskip}c@{\colskip}c}
\hline
\multirow{2}{*}{Models} & \multicolumn{2}{c@{\colskip}}{IAM} & \multicolumn{2}{c@{\colskip}}{RIMES} & \multicolumn{2}{c@{\colskip}}{IndicBAN} & \multicolumn{2}{c}{IndicDEV} \\ 
\cline{2-9}                           & QbS       & QbE        & QbS         & QbE         & QbS           & QbE          & QbS           & QbE          \\ \hline
\multirow{5}{*}{} 
                   B1             &83.12       &72.67            &86.31             &77.69             &80.37               &76.91              &81.67               &77.61              \\
                   B2             &\_       &\_            &\_             &\_             &81.04               &77.67              &82.64               &78.64              \\
                   B3             &85.1       &73.67            &87.69             &79.67             &84.67               &84.73              &85.61               &86.19              \\
                   B4             &86.94       &75.64            &90.34             &80.67             &87.67               &85.49              &88.17               &86.49              \\
                   \textbf{Ours}           &\textbf{88.69}       &\textbf{77.94}            &\textbf{92.94}             &\textbf{82.67}             &\textbf{89.34}               &\textbf{86.47}              &\textbf{90.13}               &\textbf{87.67}              \\ \hline                       
\end{tabular}
\end{table*}

\section{Experimentation Details}

\subsection{Datasets}

\begin{figure*}[]
\begin{center}
\includegraphics[width=1\linewidth]{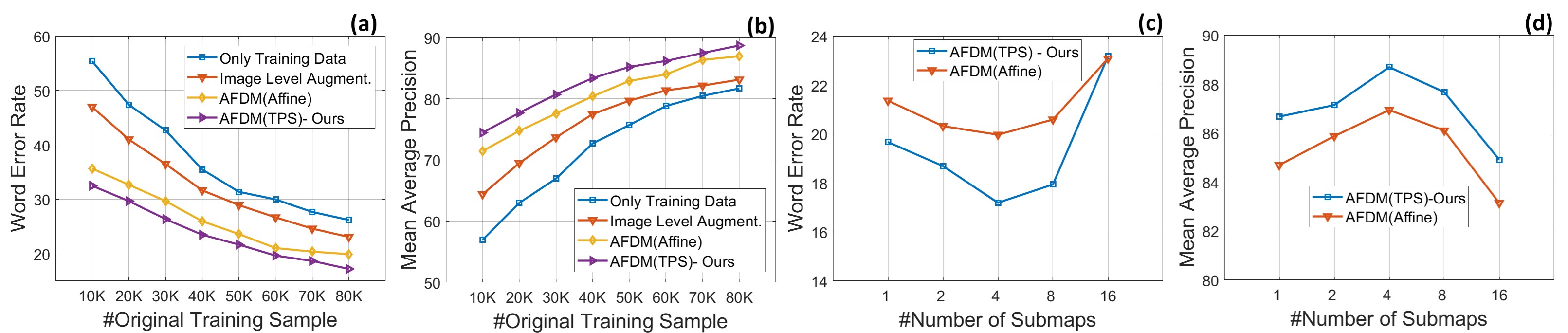}
\end{center}
   \caption{(a) Word Error Rate (WER) for HWR (unconstrained) and (b) mean Average Precision (mAP) for QbS in HWS  for different number of training samples on standard testing set using different data augmentation strategies on IAM dataset. (c) and (d) represent the performance using different sub-map partitioning schemes; the setup is described in Section 5.2.}
\label{fig:Graph1_Full}
\end{figure*}

We use two very popular datasets of Latin scripts, namely \textbf{IAM} (1,15,320 words) and \textbf{RIMES} (66,982 words) datasets, used by handwritten document image analysis community extensively. \textbf{IAM} \cite{IAMhw} is one of the largest datasets available for HWR and HWS in Latin script, allowing us to demonstrate the effectiveness of our feature warping strategy at different sizes of training sets (see Figure \ref{fig:Graph1_Full}).
In order to demonstrate the effectiveness of our model in low-resource scripts (in terms of availability of training data), we choose two Indic scripts, namely Bangla and Devanagari (Hindi), as examples to demonstrate the benefits of adversarial training via the AFDM. Hindi and Bangla are the fifth and sixth most popular languages globally \cite{pal2004indian} and use the scripts Devanagari and Bangla, respectively. Both scripts are far more complex than Latin due to the presence of modifiers\cite{roy2016hmm} and complex cursive shapes \cite{roy2016hmm} and are sparse compared to Latin \cite{grosicki2011icdar, IAMhw}. To the best of our knowledge, there exists only one publicly available dataset \cite{bhunia2018cross, roy2016hmm} which contains a total of 17,091 and 16,128 words for Bangla and Devanagari, respectively. We denote these two datasets as \textbf{IndBAN} (BANgla) and \textbf{IndDEV} (DEVanagari)  respectively. For IAM, IndBAN and IndDEV,  we use the same partition for training, validation and testing provided along with the datasets. For RIMES dataset, we follow the partition released by ICDAR 2011 competition.

\subsection{Implementation Details}

While experimenting, we notice that it is important to first pre-train the task network for a certain number of iterations so that it can learn a basic model to understand the shapes of different characters to an extent. If we start training both the networks together, we notice that the AFDM often overpowers the task network and it fails to learn meaningful representation. Therefore, first, we train the task network for 10K iterations without the AFDM. Thereafter, we include the latter to fulfill its adversarial objective of deforming the intermediate convolutional feature map. We use 500 continuous iterations to train the parameter localization network $\mathcal{A}$ alone for better initialization. It is observed that due to the large degree of flexibility TPS often finds some especially difficult deformations which task network fails to generalize later on. Hence, we use a simple trick to solve this stability issue: we only deform half of the data samples randomly in a batch through the AFDM and the rest are kept unchanged for retrieval; this greatly improves the stability issue. For the \textit{Localisation Network}, we use four convolutional layers with stride $2$ and kernel size $3\times 3$ followed by 2 fully-connected layers, finally predicting 18 parameter values using $\tanh$ activation. We keep the number of sub-map divisions ($k$) to 4. We use a batch size of 32. Following the earlier initialization, both the task-network and AFDM are trained for a total of 100K iterations alternatively. We use Adam optimizer for both task network and AFDM, however, we keep learning rate for task network to $10^{-4}$ and the same for the Localisation Network of AFDM is $10^{-3}$. PHOCNet consists of 13 convolutional layers followed an SPP layer and 3  fully connected layers and finally predicting the PHOC labels using sigmoid activation. We name these $\mathsf{conv}$-layers as follows: $\mathsf{conv1\_1}$, $\mathsf{conv1\_2}$, $\mathsf{conv2\_1}$, $\mathsf{conv2\_2}$,  $\mathsf{conv3\_1}$, $\mathsf{conv3\_2}$, $\mathsf{conv3\_3}$, $\mathsf{conv3\_4}$, $\mathsf{conv3\_5}$, $\mathsf{conv3\_6}$, $\mathsf{conv4\_1}$, $\mathsf{conv4\_2}$, $\mathsf{conv4\_3}$. There are two pooling layers($2 \times 2$) after $\mathsf{conv1\_2}$ and $\mathsf{conv2\_2}$. Every convolution layer has a kernel of size $3\times 3$ and number of filters are 64, 128, 256, and 512 for $\mathsf{conv1\_X}$, $\mathsf{conv2\_X}$, $\mathsf{conv3\_X}$, $\mathsf{conv4\_X}$ respectively. On the other hand, our CRNN framework comprises 8 $\mathsf{conv}$ layers, followed by a `Map-to-Sequence' and a 2-layer BLSTM unit. The architecture is: $\mathsf{conv1}$, $\mathsf{conv2}$, $\mathsf{conv3\_1}$, $\mathsf{conv3\_2}$,  $\mathsf{conv4\_1}$, $\mathsf{conv4\_2}$, $\mathsf{conv5\_1}$, $\mathsf{conv5\_2}$, $\mathsf{conv6}$. The first 7 layers have $3\times 3$ kernels but the last layer has a $2\times 2$ kernel. There are 64, 128 and 256 filters in $\mathsf{conv1\_X}$, $\mathsf{conv2\_X}$, $\mathsf{conv3\_X}$, and 512 filters from $\mathsf{conv4\_1}$ till $\mathsf{conv6}$; the pooling layers are after $\mathsf{conv1}$, $\mathsf{conv2}$, $\mathsf{conv3\_2}$, $\mathsf{conv4\_2}$ and $\mathsf{conv5\_2}$. While the pooling windows of the first two pool layers are $2\times 2$, the rest are $1\times 2$. Based on the experimental analysis, we introduce AFDM after $\mathsf{conv4\_1}$ layer in PHOCNet, and after $\mathsf{conv4\_1}$ layer in CRNN. It is to be noted that the input is resized to a height of 64 keeping aspect ratio same. More analysis is given in Section \ref{abla}. 

\begin{table*}[!ht]
	\caption{Mean Average Precision(mAP) on using AFDM after a specific layer in PHOCNet for Query by String.}
	\centering
	\label{table:tab4}
	\begin{tabular}{|c|c|c|c|c|c|c|c|c|c|}
	\hline
	Layers & $\mathsf{{conv3\_1}}$ & ${\mathsf{conv3\_2}}$ & ${\mathsf{conv3\_3}}$ & ${\mathsf{conv3\_4}}$ & ${\mathsf{conv3\_5}}$ & $\mathsf{{conv3\_6}}$ & $\mathsf{{conv4\_1}}$ & $\mathsf{{conv4\_2}}$ & $\mathsf{{conv4\_3}}$\\ 		\hline
	AFDM(TPS) & 85.29 & 85.20 & 85.97 & 86.94 & 87.88 & 88.19 & \textbf{88.69} & 87.81 & 87.77\\
	AFDM(Affine) & 84.13 & 84.12 & 84.53 & 85.01 & 85.33 & 85.81 & \textbf{86.94} & 86.02 & 85.24\\
\hline
	\end{tabular}
\end{table*}
\begin{SCtable*}
\label{table:tab4}
{\caption{Word Error Rate(WER) on using AFDM after a specific layer in CRNN (unconstrained).}}%
	{\begin{tabular}[\textwidth]{|c|c|c|c|c|c|c|}
	\hline
	Layers & $\mathsf{{conv3\_1}}$ & ${\mathsf{conv3\_2}}$ & ${\mathsf{conv4\_1}}$ & ${\mathsf{conv4\_2}}$ & ${\mathsf{conv5\_1}}$ & $\mathsf{{conv5\_2}}$\\ \hline
	AFDM(TPS) & 17.98 & 17.41 &\textbf{17.19}& 17.25 & 20.32 & 20.41\\
	AFDM(Affine) & 22.01 & 20.11 & \textbf{19.97}& 20.01 & 19.99 & 20.21\\
	\hline
	\end{tabular}}
	\label{table:tab4}
\end{SCtable*}

\subsection{Baseline Methods}
To the best of our knowledge, there is no prior work dealing with adversarial data augmentation strategy for HWS and HWR. Based on different popular data augmentation and transfer learning strategies, we have defined a couple of baselines to demonstrate the effectiveness of the AFDM. 
\begin{itemize}
[wide, labelwidth=!, labelindent=0pt, nosep, topsep=0pt]
\item \textbf{B1}: In this baseline, we perform different image-level data augmentation techniques mentioned in \cite{poznanski2016cnn} and \cite{sudholt2016phocnet} on the handwritten word images to increase the total number of word samples ($\sim500K$) in the training set.


\item \textbf{B2}: Here we use transfer learning strategy to alleviate the problem of data insufficiency in low resource scripts.
We train both HWR and HWS model using a large amount of data present in Latin scripts, thereafter we fix the weights till $\mathsf{conv5\_2}$($\mathsf{conv4\_2}$) layer of the CRNN(PHOCNet) network and we fine-tune rest of the layers over the available annotated data from Indic scripts.
\item \textbf{B3}: This is identical to our adversarial learning based framework, except that it deforms data in the image-space using the TPS mechanism (Section 4.2). The input to the AFDM is the original training image. 
\item \textbf{B4}: Here, we use affine transformation \cite{jaderberg2015spatial} in place of TPS, using a fewer number of parameters (six) to devise warping policies with relatively less degree of freedom for deformation. 
\end{itemize}

\subsection{Performance on HWR and HWS}

In our experiments, we use Character Error Rate (CER) and Word Error Rate (WER) metrics \cite{bluche2017scan} for HWR, while mean Average Precision (mAP) metric \cite{sudholt2016phocnet} is considered for HWS. In case of lexicon based recognition for IAM dataset, we use all the unique words present in the dataset, whereas we use lexicon provided in ICDAR 2011 competition for RIMES dataset and the lexicons provided with the original dataset are used for IndBAN and IndDEV datasets. 
From Tables 1 and 2, it is to be noted that our adversarial feature augmentation method using TPS significantly outperforms \textbf{B1} which uses different image level data augmentation techniques as seen in \cite{poznanski2016cnn, shi2017end} together. This signifies that only image level ``handcrafted" data augmentation cannot improve the performance significantly even if we increase the the number of data-samples through possible transformations. 
We notice that weight initialization from pretrained weights in \textbf{B2} helps to increase the performance for both HWR and HWS to a reasonable extent and also speeds up the training procedure significantly. Both \textbf{B3} and \textbf{B4} are adversarial frameworks. From the results on both HWR and HWS, it can be concluded that adversarial data augmentation works better while introduced in the intermediate layers of the convolutional network rather than adversarial deformation in image space as done in \textbf{B3}. Also, TPS performs better than simple affine transformation in \textbf{B4} due to greater degree of flexibility in deformation. 

Overall, the improvement due to adversarial data augmentation is clearly higher for both IndBAN and IndDEV. Also, performance is better in IndBAN and IndDEV dataset than other two datasets inspite of our claim of having more complexity in Bangla and Devanagari script. The major reason behind this is that IndBAN and IndDEV datasets have multiple copies of same words by same author in both training and testing sets as well as simpler words (having 4 characters on average), while the IAM dataset has more complex samples in testing sets. Word retrieval in real-world scenarios of Bangla and Devanagari script is far more complex than what it is in unseen testing set. Moreover, due to limited training data as well as a large number of character classes \cite{bhunia2018cross, roy2016hmm}, image level data augmentation can not generalize the model well in testing set, giving poor performance for both HWR and HWS.  In contrast, the proposed method using adversarial learning helps in significant performance gain compared to image level data augmentation.

\subsection{Ablation Study}\label{abla}
We have comprehensively studied the improvements achieved from different augmentation techniques at various training data sizes on the IAM dataset. We experiment over 8 instances with our training set size ranging from 10K to 80K, using the standard testing set for evaluation. 
From Figure \ref{fig:Graph1_Full}, it is evident that the proposed method performs well in the low-data regime, producing a reasonable improvement over image-level augmentation. It is to be noted that with increasing training data, the improvement gained by our model over other baselines (which do not use adversarial augmentation) gets reduced. 
We also evaluated the performance by including the AFDM at different positions of the ConvNet feature extraction units in CRNN and PHOCNet (shown in Table 3 and 4). We observe that if the AFDM is inserted between shallower layers, the model diverges and we do not achieve a desirable result. Better performance along with improved stability in training is observed in the mid-to-deeper parts of the task network which encode a higher-level understanding of the extracted feature information. The performance again drops at very deep layers. We also evaluate the performance of the model by partitioning the original feature map into 1, 2, 4, 8 and 16 sub-maps using rest of the standard setup. It was noticed that 4 divisions provide the optimum result (Figure \ref{fig:Graph1_Full}).  

\paragraph{Adversarial \vs Non-adversarial Learning:}
In contrast to the AFDM that is based on STN\cite{jaderberg2015spatial} and trained using adversarial objective, an alternative (\textit{non-adversarial}) could be the work by Shi \etal\cite{shi2016robust} where STN is used to rectify the spatial orientation of a word-image to make recognition easier for Sequence Recognition Network \cite{shi2016robust} according to the original philosophy of \cite{jaderberg2015spatial}. Following \cite{shi2016robust}, we introduce an STN module with TPS before the CRNN and PHOCNet architecture and train the complete architecture (STN + Task Network) in an end-to-end manner with the task loss objective (eqn. \ref{loss_1}), keeping the rest of the standard experimental setup of task network same. The unconstrained WER for non-adversarial pipeline using STN is 20.07\% and the mAP value for QbS is 85.64\%, trailing behind the proposed adversarial framework by 3.51\%(WER) and 3.05\%(mAP) respectively. Next, we divide the IAM dataset into hard and easy word samples using the framework of Mor \etal \cite{mor2018confidence} with CRNN as baseline recognizer. 
We consider top 70\% word images as easy samples and 30\% as hard samples based on the confidence score. High score signifies easily recognizable images without much deformation, while images with lower scores contain ample deformation in them. We train both the adversarial and non-adversarial pipeline using these easy samples and test on hard samples. This experimental setup challenges the models to learn invariance that can generalize for hard unseen word samples which are absent during training. It is observed that while non-adversarial pipeline provides 40.22\% unconstrained WER (71.31 mAP-QbS), our adversarial framework achieves 27.64\% WER (82.67 mAP-QbS) outperforming the non-adversarial alternatives by a large margin of 12.58\% WER (11.36 mAP-QbS). Although the objective of both of these pipelines is to learn a robust model invariant to different types of deformation in handwritten data, the non-adversarial method tries to learn the invariance only from available training data while failing to generalize on unseen irregularities and deformations. Due to free-flow nature of handwriting, it is not possible to include every possible variation in the training dataset. Hence, our adversarial framework proves useful to learn a robust model that can generalize well on unseen deformations which are absent in sparse datasets.
 
\section{Conclusion}
We study a common difficulty often faced by researchers exploring handwriting recognition in low-resource scripts and try to overcome the limitations of generic data augmentation strategies. The AFDM can be flexibly added to frameworks for both word-spotting and recognition, allowing deep networks to generalize well even in low-data settings.  Rather than augmenting handwritten data in image space using ``handcrafted" techniques, adversarially warping the intermediate feature-space using TPS is a scalable solution to overcome the dearth of variations seen in some sparse training datasets. The higher degree of flexibility incorporated by TPS with the adversarial parameterisation strategy goes a long way to incorporate rare unseen variations, beating deformation policies that frameworks can easily overfit to.

{\small
\bibliographystyle{ieee}
\bibliography{egbib}
}

\end{document}